\documentclass[11pt,a4paper]{article}
\usepackage[hyperref]{style/acl2020}
\usepackage{times}
\usepackage{latexsym}

\usepackage{microtype}

\aclfinalcopy 
\def\aclpaperid{260} 

\usepackage{url}
\usepackage[utf8]{inputenc}

\usepackage{graphicx}
\usepackage{amsfonts}
\usepackage{mdwlist}
\usepackage{multirow}
\usepackage{algorithm}
\usepackage[noend]{algpseudocode}
\usepackage{bm}
\usepackage{physics}
\usepackage{caption}
\usepackage{subcaption}
\usepackage{amsmath}
\usepackage{mathtools}

\newif\ifcomment\commenttrue

\usepackage{framed}
\usepackage{mdwlist}
\usepackage{latexsym}
\usepackage{colortbl}
\usepackage{xcolor}
\usepackage{nicefrac}
\usepackage{booktabs}
\usepackage{amsfonts}
\usepackage[T1]{fontenc}
\usepackage{bold-extra}
\usepackage{amsmath}
\usepackage{amssymb}
\usepackage{bm}
\usepackage{graphicx}
\usepackage{mathtools}
\usepackage{microtype}
\usepackage{multirow}
\usepackage{multicol}
\usepackage{latexsym,comment}
\usepackage[normalem]{ulem}

\newcommand*{\missingreference}{{\Huge \colorbox{red}{?reference?}}}
\newcommand*{\missingcitation}{{\Huge \colorbox{red}{?citation?}}}
\makeatletter
\def\@setref#1#2#3{%
  \ifx#1\relax
    \protect\G@refundefinedtrue
    \nfss@text{\reset@font\missingreference}%
    \@latex@warning{Reference `#3' on page \thepage \space
      undefined}%
  \else
    \expandafter#2#1\null
  \fi}
\def\@citex[#1]#2{\leavevmode
  \let\@citea\@empty
  \@cite{\@for\@citeb:=#2\do
    {\@citea\def\@citea{,\penalty\@m\ }%
      \edef\@citeb{\expandafter\@firstofone\@citeb\@empty}%
      \if@filesw\immediate\write\@auxout{\string\citation{\@citeb}}\fi
      \@ifundefined{b@\@citeb}{\hbox{\reset@font\missingcitation}%
        \G@refundefinedtrue
        \@latex@warning
        {Citation `\@citeb' on page \thepage \space undefined}}%
      {\@cite@ofmt{\csname b@\@citeb\endcsname}}}}{#1}}
\makeatother

\newcommand{\gem}[1]{\mbox{\textsc{gem}}}
\newcommand{\abr}[1]{\textsc{#1}}

\newcommand{\emaillink}[1]{ {\small \href{mailto://#1}{\texttt{#1}}}}

\newcommand{\hidetext}[1]{}
\newcommand{\ignore}[1]{}

\ifcomment
  \newcommand{\pinaforecomment}[3]{\colorbox{#1}{\parbox{.8\linewidth}{#2: #3}}}
\else
  \newcommand{\pinaforecomment}[3]{}
\fi

\newcommand{\smallurl}[1]{ \begin{tiny}\url{#1}\end{tiny}}

\definecolor{lightblue}{HTML}{3cc7ea}
\definecolor{CUgold}{HTML}{CFB87C}
\definecolor{grey}{rgb}{0.95,0.95,0.95}
\definecolor{ceil}{rgb}{0.57, 0.63, 0.81}
\definecolor{UMDred}{HTML}{ed1c24}
\definecolor{UMDyellow}{HTML}{ffc20e}

\ifcomment
\newcommand{\mzcomment}[1]{ \colorbox{lightblue}{   \parbox{.8\linewidth}{ MZ: #1}  }}
\else
\newcommand{\mzcomment}[1]{}
\fi

\newcommand{\vect}[1]{\bm{\mathbf{#1}}}

\newcommand{\figfile}[1]{2020_acl_refine_clwe/figures/#1}
\newcommand{\autofig}[1]{2020_acl_refine_clwe/auto_fig/#1}

\title{Why Overfitting Isn't Always Bad: \\
Retrofitting Cross-Lingual Word Embeddings to Dictionaries}

\author{Mozhi Zhang\thanks{$^\star$Equal contribution} \\
  \textsc{cs} and \textsc{umiacs} \\
  University of Maryland \\
  \emaillink{mozhi@cs.umd.edu} \\\And
  Yoshinari Fujinuma\footnotemark[1] \\
  Computer Science \\
  University of Colorado \\
  \emaillink{fujinumay@gmail.com} \\\AND
  Michael J. Paul \\
  Information Science \\
  University of Colorado \\
  \emaillink{mpaul@colorado.edu} \\\And
  Jordan Boyd-Graber \\
  \abr{umd} \textsc{cs}, iSchool, \textsc{umiacs}, and \textsc{lsc}  \\
  and Google Research Z\"urich \\
  \emaillink{jbg@boydgraber.org} \\}

\date{}

\begin{document}
\maketitle
\begin{abstract}

Cross-lingual word embeddings~(\abr{clwe}) are often evaluated on bilingual
lexicon induction~(\abr{bli}).
Recent \abr{clwe} methods use linear projections, which underfit the training
dictionary, to generalize on \abr{bli}.
However, underfitting can hinder generalization to other downstream tasks that 
rely on words from the training dictionary.
We address this limitation by \emph{retrofitting} \abr{clwe} to the training
dictionary, which pulls training translation pairs closer in the
embedding space and overfits the training dictionary.
This simple post-processing step often improves accuracy on two downstream
tasks, despite lowering \abr{bli} test accuracy.
We also retrofit to both the training dictionary and a
synthetic dictionary induced from \abr{clwe}, which sometimes generalizes
even better on downstream tasks.
Our results confirm the importance of fully exploiting the training dictionary
in downstream tasks and explains why \abr{bli} is a flawed CLWE evaluation.

\end{abstract}

\section{Introduction}

Cross-lingual word embeddings~(\abr{clwe}) map words across languages to
a shared vector space.
Recent supervised \abr{clwe} methods follow a projection-based
pipeline~\citep{mikolov-13b}.
Using a training dictionary, a linear projection maps
pre-trained monolingual embeddings to a multilingual space.
While \abr{clwe} enable many multilingual
tasks~\citep{klementiev-12,guo-15,zhang-16,ni-17},
most recent work only evaluates \abr{clwe} on bilingual lexicon
induction~(\abr{bli}).
Specifically, a set of test words are translated with a retrieval heuristic
(e.g., nearest neighbor search) and compared against gold translations.
\abr{bli} accuracy is easy to compute and captures the desired property of
\abr{clwe} that translation pairs should be close.
However, \abr{bli} accuracy does not always correlate with accuracy on
downstream tasks such as cross-lingual document
classification and dependency parsing~\citep{ammar-16,fujinuma-19,glavas-19}.

\begin{figure}[t!]
\centering
\includegraphics[width=\linewidth]{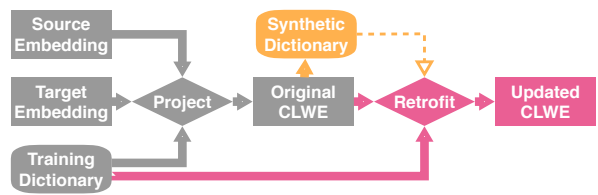}
\caption{\label{fig:arch}
  To fully exploit the training dictionary, we retrofit projection-based
  \abr{clwe} to the training dictionary as a post-processing step (pink
  parts).
  To preserve correctly aligned translations in the original \abr{clwe}, we
  optionally retrofit to a synthetic dictionary induced from the original
  \abr{clwe} (orange parts).
}
\end{figure}

Let's think about why that might be.
\abr{bli} accuracy is only computed on \emph{test} words.
Consequently, \abr{bli} hides linear projection's inability to
align all training translation pairs at once; i.e., projection-based
\abr{clwe} \emph{underfit} the training dictionary.
Underfitting does not hurt \abr{bli} test accuracy,
because test words are excluded from the training dictionary in \abr{bli}
benchmarks.
However, words from the training dictionary may be nonetheless predictive in
downstream tasks; e.g., if ``good'' is in the training dictionary, knowing its translation is useful for
multilingual sentiment analysis.

In contrast, \emph{overfitting} the training dictionary hurts \abr{bli} but can
improve downstream models.
We show this by adding a simple post-processing step to
projection-based pipelines (Figure~\ref{fig:arch}).
After training supervised \abr{clwe} with a projection, we \emph{retrofit}~\citep{faruqui-15} the \abr{clwe} to the same
training dictionary.
This step pulls training translation pairs closer and overfits:
the updated embeddings have perfect \abr{bli} training accuracy, but \abr{bli}
\emph{test} accuracy drops.
Empirically, retrofitting improves accuracy in two downstream tasks other than
\abr{bli}, confirming the importance of fully exploiting the training
dictionary.

Unfortunately, retrofitting to the training dictionary may inadvertently push
some translation pairs further away.
To balance between fitting the training dictionary and generalizing on other
words, we explore retrofitting to both the training dictionary and a
\emph{synthetic dictionary} induced from the \abr{clwe}.
Adding the synthetic dictionary keeps some correctly aligned translations in
the original \abr{clwe} and can further improve downstream models
by striking a balance between training and test \abr{bli} accuracy.

In summary, our contributions are two-fold.
First, we explain \emph{why} \abr{bli} does not reflect downstream task
accuracy.
Second, we introduce two post-processing methods to improve downstream models
by fitting the training dictionary better.

\section{Limitation of Projection-Based \abr{clwe}}
\label{sec:clwe}

This section reviews projection-based \abr{clwe}.
We then discuss how \abr{bli} evaluation obscures the limitation of
projection-based methods.

Let $\vect{X}\in\mathbb{R}^{d\cross n}$ be a pre-trained $d$-dimensional word
embedding matrix for a source language, where each column
$\vect{x}_i\in\mathbb{R}^d$ is the vector for word $i$ from the source language
with vocabulary size $n$, and let $\vect{Z}\in\mathbb{R}^{d\cross m}$ be a
pre-trained word embedding matrix for a target language with vocabulary size $m$.
Projection-based \abr{clwe} maps $\vect{X}$ and $\vect{Z}$ to a shared
space.
We focus on supervised methods that learn the projection from a training
dictionary~$\mathcal{D}$ with translation pairs~$(i, j)$.

\citet{mikolov-13b} first propose projection-based \abr{clwe}.
They learn a linear projection $\vect{W}\in\mathbb{R}^{d\cross d}$ from
$\vect{X}$ to $\vect{Z}$ by minimizing distances between translation pairs in
a training dictionary:
\begin{equation}
\min_{\vect{W}} \sum_{(i,j) \in \mathcal{D}} \| \vect{W} \vect{x}_i - \vect{z}_j \|_2^2.
\end{equation}
Recent work improves this method with different optimization
objectives~\citep{dinu-15,joulin-18},
orthogonal constraints on $\vect{W}$~\citep{xing-15,artetxe-16,smith-17},
pre-processing~\citep{zhang-19},
and subword features~\citep{chaudhary-18,czarnowska-19,zhang-20}.

Projection-based methods \emph{underfit}---a linear projection has limited
expressiveness and cannot perfectly align all training pairs.
Unfortunately, this weakness is not transparent when using \abr{bli} as the
standard evaluation for \abr{clwe}, because \abr{bli} test sets omit training
dictionary words.
However, when the training dictionary covers words that help downstream
tasks, underfitting limits generalization to other tasks.
Some \abr{bli} benchmarks use frequent words for training and infrequent words
for testing~\citep{mikolov-13b,conneau-18}.
This mismatch often appears in real-world data, because frequent
words are easier to find in digital dicitonaries~\citep{czarnowska-19}.
Therefore, training dictionary words are often more important in downstream
tasks than test words.

\begin{figure*}[t]
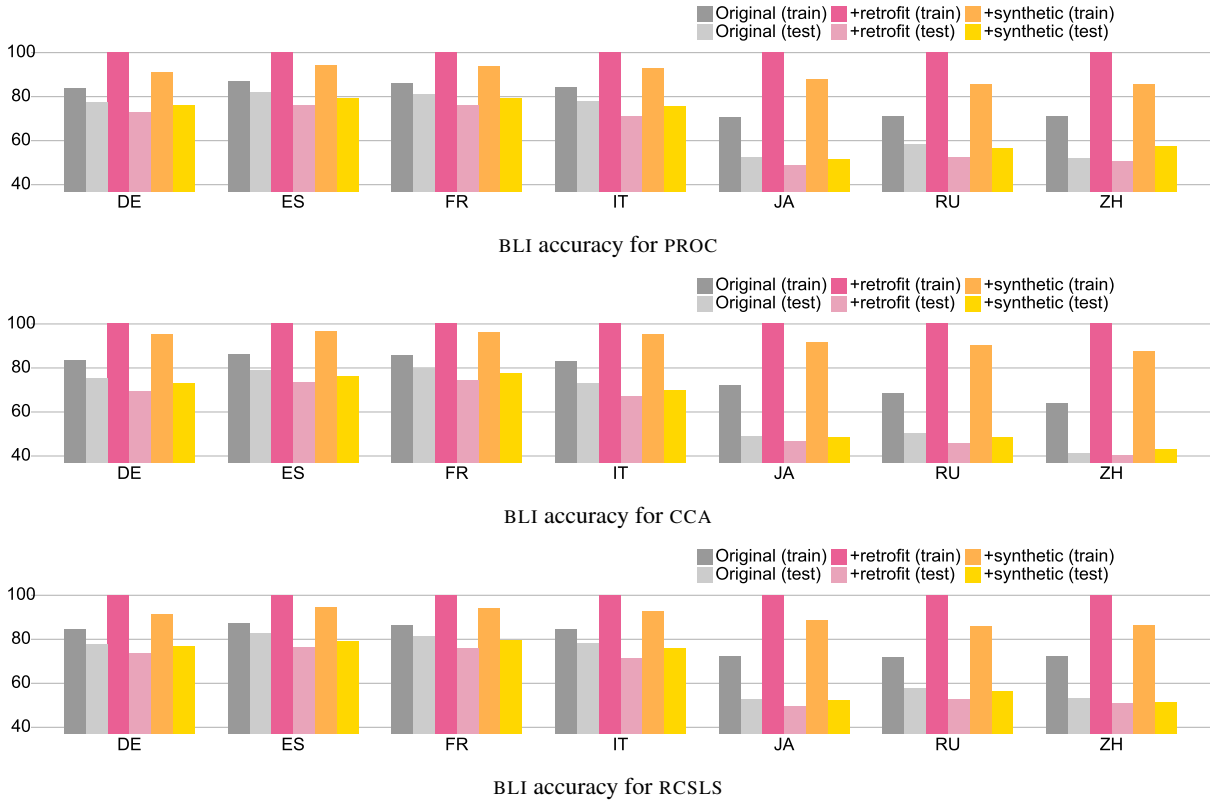

  \centering
  \begin{subfigure}{\linewidth}
    \centering
    \includegraphics[width=\linewidth]{\autofig{bli-proc}}
    \caption*{\abr{bli} accuracy for \abr{proc}}
  \end{subfigure}
  \begin{subfigure}{\linewidth}
    \centering
    \includegraphics[width=\linewidth]{\autofig{bli-cca}}
    \caption*{\abr{bli} accuracy for \abr{cca}}
  \end{subfigure}
  \begin{subfigure}{\linewidth}
    \centering
    \includegraphics[width=\linewidth]{\autofig{bli-rcsls}}
    \caption*{\abr{bli} accuracy for \abr{rcsls}}
  \end{subfigure}
  \caption{Train and test accuracy (P@1) for \abr{bli} on \abr{muse};
  Projection-based \abr{clwe} underfit the training dictionary (gray),
  but retrofitting to the training dictionary overfits (pink).
  Adding a synthetic dictionary balances between training and test accuracy
  (orange).}
  \label{fig:bli}
\end{figure*}

\section{Retrofitting to Dictionaries}

To fully exploit the training dictionary, we explore a simple post-processing
step that \emph{overfits} the dictionary: we first train projection-based
\abr{clwe} and then \emph{retrofit} to the training
dictionary~(pink parts in Figure~\ref{fig:arch}).
Retrofitting was originally introduced for refining monolingual word embeddings
with synonym constraints from a lexical ontology~\citep{faruqui-15}.
For \abr{clwe}, we retrofit using the training dictionary $\mathcal{D}$ as the
ontology.

Intuitively, retrofitting pulls translation pairs closer while minimizing
deviation from the original \abr{clwe}.
Let $\vect{X}'$ and $\vect{Z}'$ be \abr{clwe} trained by a projection-based
method,
where $\vect{X}'=\vect{W}\vect{X}$ are the projected source embeddings and
$\vect{Z}'=\vect{Z}$ are the target embeddings.
We learn new \abr{clwe} $\hat{\vect{X}}$ and $\hat{\vect{Z}}$ by minimizing
\begin{equation}
  L = L_a + L_b,
\end{equation}
where $L_a$ is the squared distance between the updated \abr{clwe} from the
original \abr{clwe}:
\begin{equation}
  L_a = \alpha \|\hat{\vect{X}}-\vect{X}'\|^2 + \alpha \|\hat{\vect{Z}} - \vect{Z}'\|^2,
\end{equation}
and $L_b$ is the total squared distance between translations in the dictionary:
\begin{equation}
  L_b = \sum_{(i,j)\in\mathcal{D}} \beta_{ij} \|\hat{\vect{x}}_i - \hat{\vect{z}}_j\|^2.
\end{equation}
We use the same $\alpha$ and $\beta$ as \citet{faruqui-15} to balance the two objectives.

Retrofitting tends to overfit.
If $\alpha$ is zero, minimizing $L_b$ collapses each training pair
to the same vector.
Thus, all training pairs are perfectly aligned.
In practice, we use a non-zero $\alpha$ for regularization, but the updated
\abr{clwe} still have perfect training \abr{bli}
accuracy~(Figure~\ref{fig:bli}).
If the training dictionary covers predictive words, we expect retrofitting to
improve downstream task accuracy.

\subsection{Retrofitting to Synthetic Dictionary}

While retrofitting brings pairs in the training dictionary closer,
the updates may also separate translation pairs outside of the dictionary
because retrofitting ignores words outside the training dictionary.
This can hurt both \abr{bli} test accuracy and downstream task accuracy.
In contrast, projection-based methods underfit but can discover translation
pairs outside the training dictionary.
To keep the original \abr{clwe}'s correct translations, we
retrofit to both the training dictionary and a \emph{synthetic dictionary}
induced from \abr{clwe}~(orange, Figure~\ref{fig:arch}).

Early work induces dictionaries from \abr{clwe} through
nearest-neighbor search~\citep{mikolov-13b}.  We instead use
cross-domain similarity local scaling~\citep[\abr{csls}]{conneau-18},
a translation heuristic more robust to hubs~\citep{dinu-15} (a word is the
nearest neighbor of many words).
We build a synthetic dictionary $\mathcal{D'}$ with word pairs that are
\emph{mutual} \abr{csls} nearest neighbors.
We then retrofit the \abr{clwe} to a combined dictionary
$\mathcal{D}\cup\mathcal{D}'$.
The synthetic dictionary keeps closely aligned word pairs in the original
\abr{clwe}, which sometimes improves downstream models.

\begin{figure*}
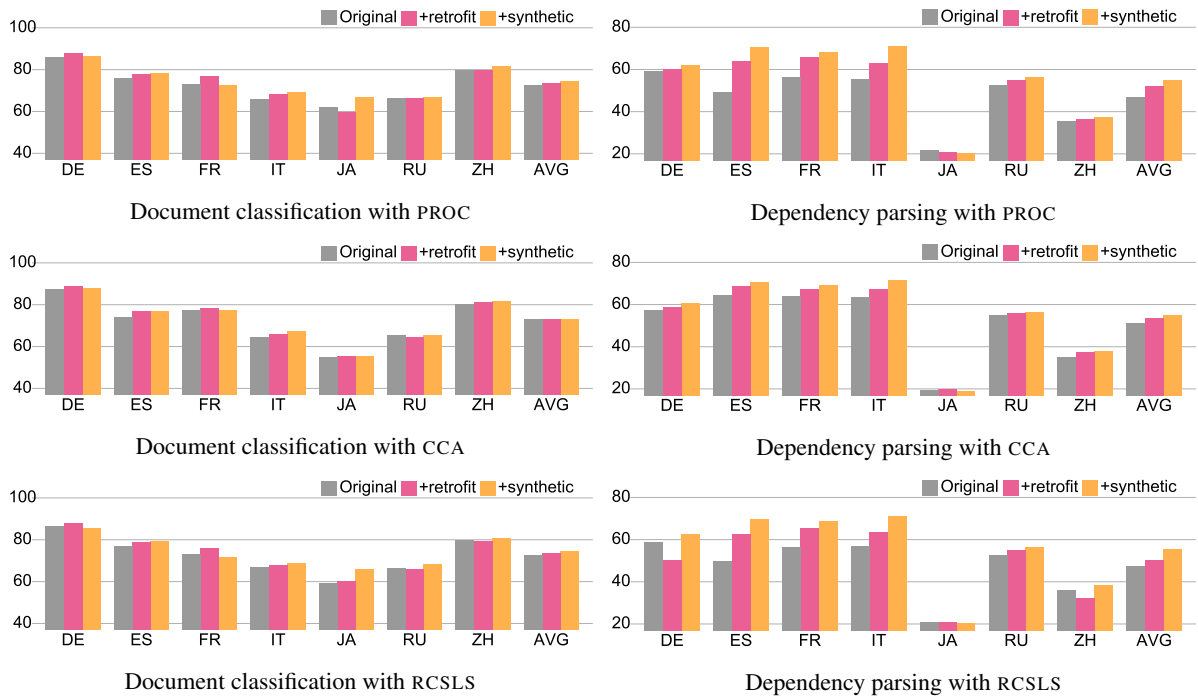

  \centering
  \begin{subfigure}{.49\linewidth}
    \centering
    \includegraphics[width=\linewidth]{\autofig{cldc-proc}}
    \caption*{Document classification with \abr{proc}}
  \end{subfigure}
  \begin{subfigure}{.49\linewidth}
    \centering
    \includegraphics[width=\linewidth]{\autofig{cldp-proc}}
    \caption*{Dependency parsing with \abr{proc}}
  \end{subfigure}
  \begin{subfigure}{.49\linewidth}
    \centering
    \includegraphics[width=\linewidth]{\autofig{cldc-cca}}
    \caption*{Document classification with \abr{cca}}
  \end{subfigure}
  \begin{subfigure}{.49\linewidth}
    \centering
    \includegraphics[width=\linewidth]{\autofig{cldp-cca}}
    \caption*{Dependency parsing with \abr{cca}}
  \end{subfigure}
  \begin{subfigure}{.49\linewidth}
    \centering
    \includegraphics[width=\linewidth]{\autofig{cldc-rcsls}}
    \caption*{Document classification with \abr{rcsls}}
  \end{subfigure}
  \begin{subfigure}{.49\linewidth}
    \centering
    \includegraphics[width=\linewidth]{\autofig{cldp-rcsls}}
    \caption*{Dependency parsing with \abr{rcsls}}
  \end{subfigure}
  \caption{For each \abr{clwe}, we
  report accuracy for document classification (left) and unlabeled attachment
  score (\abr{uas}) for dependency parsing (right).
  Compared to the original embeddings (gray), retrofitting to the training
  dictionary (pink) improves average downstream task scores, confirming that
  fully exploiting the training dictionary helps downstream tasks.
  Adding a synthetic dictionary (orange) further improves test accuracy in some
  languages.}
  \label{fig:downstream}
\end{figure*}

\section{Experiments}

We retrofit three projection-based \abr{clwe} to their training dictionaries
and synthetic dictionaries.\footnote{Code at \smallurl{https://go.umd.edu/retro_clwe}.}
We evaluate on \abr{bli} and two downstream tasks.
While retrofitting decreases test \abr{bli} accuracy, it often improves downstream
models.

\subsection{Embeddings and Dictionaries}

We align English embeddings with six target languages: German~(\abr{de}),
Spanish~(\abr{es}), French~(\abr{fr}), Italian~(\abr{it}), Japanese~(\abr{ja}),
and Chinese~(\abr{zh}).
We use 300-dimensional fastText vectors trained on Wikipedia and Common
Crawl~\citep{grave-18}. 
We lowercase all words, only keep the 200K most frequent words, and apply five
rounds of Iterative Normalization~\citep{zhang-19}.

We use dictionaries from \abr{muse}~\citep{conneau-18}, a popular \abr{bli}
benchmark, with standard splits: train on 5K source word translations and test
on 1.5K words for \abr{bli}.
For each language, we train three projection-based \abr{clwe}:
canonical correlation analysis~\citep[\abr{cca}]{faruqui-14}, Procrustes
analysis~\citep[\abr{proc}]{conneau-18}, and Relaxed \abr{csls}
loss~\citep[\abr{rcsls}]{joulin-18}.
We retrofit these \abr{clwe} to the training dictionary (pink in figures) and
to both the training and the synthetic dictionary (orange in figures).

In \abr{muse}, words from the training dictionary have higher
frequencies than words from the test
set.\footnote{\url{https://github.com/facebookresearch/MUSE/issues/24}}
For example, the most frequent word in the English-French test
dictionary is ``torpedo'', while the training dictionary has
translations for frequent words such as ``the'' and ``good''.  As
discussed in \S\ref{sec:clwe}, more frequent words are likely to
be more salient in downstream tasks, so underfitting these more
frequent training pairs hurts generalization to downstream
tasks.\footnote{A pilot study confirms that retrofitting to infrequent
  word pairs is less effective.}

\subsection{Intrinsic Evaluation: \abr{bli}}

We first compare \abr{bli} accuracy on both training and test
dictionaries~(Figure~\ref{fig:bli}).
We use \abr{csls} to translate words with default parameters.
The original projection-based \abr{clwe} have the highest test accuracy but
underfit the training dictionary.
Retrofitting to the training dictionary perfectly fits the training
dictionary but drops test accuracy.
Retrofitting to the combined dictionary splits the difference: higher test
accuracy but lower train accuracy.
These three modes offer a continuum between \abr{bli} test and
training accuracy.

\subsection{Extrinsic Evaluation: Downstream Tasks}

We compare \abr{clwe} on two downstream tasks: document classification and
dependency parsing.
We fix the embeddng layer of the model to \abr{clwe} and use the zero-shot
setting, where a model is trained in English and evaluated in the target
language.

\paragraph{Document Classification}
Our first downstream task is document-level classification.
We use MLDoc, a multilingual classification benchmark~\citep{schwenk-18} using the standard split with 1K training and 4K test documents.
Following \citet{glavas-19}, we use a convolutional neural network~\cite{kim-14}.
We apply $0.5$ dropout to the final layer, run Adam~\citep{kingma-15} with default parameters for ten epochs, and
report the average accuracy of ten runs.

\paragraph{Dependency Parsing}
We also test on dependency parsing, a structured prediction task. 
We use Universal Dependencies~\citep[v2.4]{ud2.4} with the standard split.
We use the biaffine parser~\citep{Dozat2017biaffine} in 
AllenNLP~\citep{Gardner2017AllenNLP} with the same hyperparameters as
\citet{ahmad-19}.
To focus on the influence of \abr{clwe}, we remove part-of-speech features~\citep{ammar-16}.
We report the average unlabeled attachment score (\abr{uas}) of five runs.

\paragraph{Results}
Although training dictionary retrofitting lowers \abr{bli} test
accuracy, it improves both downstream tasks' test accuracy (Figure~\ref{fig:downstream}).
This confirms that over-optimizing the test \abr{bli} accuracy can hurt
downstream tasks because training dictionary words are also important.
The synthetic dictionary further improves downstream models,
showing that generalization to downstream tasks must balance
between \abr{bli} training and test accuracy.

\paragraph{Qualitative Analysis}
As a qualitative example, coordinations improve after
retrofitting to the training dictionary.
For example, in the German sentence ``Das Lokal ist sauber, hat einen
gem\"utlichen `Raucherraum' und wird gut besucht'', the bar (``Das Lokal'') has
three properties: it is clean, has a smoking room, and is popular.
However, without retrofitting, the final property ``besucht'' is
connected to ``hat'' instead of ``sauber''; i.e., the final clause
stands on its own.
After retrofitting to the English-German training dictionary, ``besucht'' is
moved closer to its English translation ``visited'' and is correctly parsed as
a property of the bar.

\section{Related Work}

Previous work proposes variants of retrofitting broadly called \emph{semantic
specialization} methods.
Our pilot experiments found similar trends when replacing retrofitting with
Counter-fitting~\citep{mrksic-16} and Attract-Repel~\citep{mrksic-17}, so we
focus on retrofitting. 

Recent work applies semantic specialization to \abr{clwe} by using multilingual
ontologies~\citep{mrksic-17}, transferring a monolingual ontology across
languages~\citep{ponti-19}, and asking bilingual speakers to annotate
task-specific keywords~\citep{yuan-19b}.
We instead re-use the training dictionary of the \abr{clwe}.

Synthetic dictionaries are previously used to iteratively refine a linear
projection~\citep{artetxe-17,conneau-18}.
These methods still underfit because of the linear constraint.
We instead retrofit to the synthetic dictionary to fit the training dictionary
better while keeping some generalization power of projection-based \abr{clwe}.

Recent work investigates cross-lingual contextualized embeddings as an
alternative to
\abr{clwe}~\citep{eisenschlos-19,lample-19,huang-19,wu-19,conneau-20}.
Our method may be applicable, as recent work also applies projections to
contextualized embeddings~\citep{aldarmaki-19,schuster-19,wang-20,wu-20}.

\section{Conclusion and Discussion}

Popular \abr{clwe} methods are optimized for \abr{bli} test accuracy.
They underfit the training dictionary, which hurts downstream models.
We use retrofitting to fully exploit the training dictionary.
This post-processing step improves downstream task accuracy despite lowering
\abr{bli} test accuracy.
We then add a synthetic dictionary to balance \abr{bli} test and training
accuracy, which further helps downstream models on average.

\abr{bli} test accuracy does not always correlate with downstream task
accuracy because words from the training dictionary are ignored.
An obvious fix is adding training words to the \abr{bli} test set.
However, it is unclear how to balance between training and test words.
\abr{bli} accuracy assumes that all test words are equally important,
but the importance of a word depends on the downstream task; e.g., ``the'' is
irrelevant in document classification but important in dependency parsing.
Therefore, future work should focus on downstream tasks instead of \abr{bli}.

We focus on retrofitting due to its simplicity.
There are other ways to fit the dictionary better; e.g., using a non-linear
projection such as a neural network.
We leave the exploration of non-linear projections to future work.

\section*{Acknowledgement}

This research is supported by NSF grant IIS-1564275 and by ODNI, IARPA, via the
BETTER Program contract \#2019-19051600005.
The views and conclusions contained herein are those of the authors and should
not be interpreted as necessarily representing the official policies, either
expressed or implied, of ODNI, IARPA, or the U.S. Government. The U.S.
Government is authorized to reproduce and distribute reprints for governmental
purposes notwithstanding any copyright annotation therein.

\clearpage

\bibliography{bib/journal-full,bib/yoshinari,bib/mozhi,bib/jbg}
\bibliographystyle{style/acl_natbib}

\end{document}